\documentclass[10pt,twocolumn,letterpaper]{article}

\usepackage{wacv}      

\usepackage{graphicx}
\usepackage{amsmath}
\usepackage{amssymb}
\usepackage{booktabs}

\usepackage{amsmath,amsfonts,bm}

\def\eqref#1{equation~\ref{#1}}

\def\1{\bm{1}}

\DeclareMathAlphabet{\mathsfit}{\encodingdefault}{\sfdefault}{m}{sl}
\SetMathAlphabet{\mathsfit}{bold}{\encodingdefault}{\sfdefault}{bx}{n}

\DeclareMathOperator*{\argmax}{arg\,max}

\usepackage[accsupp]{axessibility}
\usepackage{wrapfig}
\usepackage{lipsum}
\usepackage{amsmath}
\usepackage{amssymb}
\usepackage{comment}
\usepackage{graphicx}
\usepackage{rotating}
\usepackage{mathtools}
\usepackage{balance}
\usepackage{subcaption}
\usepackage{caption}
\usepackage{relsize}
\usepackage{multirow}
\usepackage{adjustbox}
\usepackage{nccmath}
\usepackage{epsfig}
\usepackage{amsthm,mathrsfs,bm}
\usepackage{dsfont,color,bbm}
\usepackage{algorithm}
\usepackage[noend]{algorithmic}
\renewcommand{\algorithmiccomment}[1]{\bgroup\hfill\tiny//~#1\egroup}
\usepackage{cite,epstopdf,epsf}
\usepackage{resizegather}
\usepackage{booktabs}
\usepackage{float}
\usepackage{xr}
\usepackage{enumitem}

\DeclarePairedDelimiter{\nfloor}\lfloor\rfloor

\usepackage[pagebackref,breaklinks,colorlinks]{hyperref}

\usepackage[capitalize]{cleveref}
\crefname{section}{Sec.}{Secs.}
\Crefname{section}{Section}{Sections}
\Crefname{table}{Table}{Tables}
\crefname{table}{Tab.}{Tabs.}

\DeclareRobustCommand*{\IEEEauthorrefmark}[1]{\raisebox{0pt}[0pt][0pt]{\textsuperscript{\footnotesize\ensuremath{\ifcase#1\or *\or \dagger\or \ddagger\or%
    \mathsection\or \mathparagraph\or \|\or **\or \dagger\dagger%
    \or \ddagger\ddagger \else\textsuperscript{\expandafter\romannumeral#1}\fi}}}}

\begin{document}

\title{Adaptive Deep Neural Network Inference Optimization with EENet}

\author{
Fatih Ilhan\IEEEauthorrefmark{1}, 
Ka-Ho Chow\IEEEauthorrefmark{1}, 
Sihao Hu\IEEEauthorrefmark{1}, 
Tiansheng Huang\IEEEauthorrefmark{1}, 
Selim Tekin\IEEEauthorrefmark{1}, 
Wenqi Wei\IEEEauthorrefmark{1}, 
Yanzhao Wu\IEEEauthorrefmark{1}, 
\\ 
Myungjin Lee\IEEEauthorrefmark{2}, 
Ramana Kompella\IEEEauthorrefmark{2}, 
Hugo Latapie\IEEEauthorrefmark{2}, 
Gaowen Liu\IEEEauthorrefmark{2}, 
Ling Liu\IEEEauthorrefmark{1} 
\\
\IEEEauthorrefmark{1}Georgia Institute of Technology, Atlanta, GA, USA \quad \IEEEauthorrefmark{2}CISCO Research, San Jose, CA, USA
\\
{\tt\small\IEEEauthorrefmark{1}\{filhan, khchow, sihaohu, thuang, stekin6, wenqiwei, yanzhaowu, ling.liu\}@gatech.edu}, 
\\
{\tt\small\IEEEauthorrefmark{2}\{myungjle, rkompell, hlatapie, gaoliu\}@cisco.com}
}

\maketitle

\begin{abstract}
Well-trained deep neural networks (DNNs) treat all test samples equally during prediction. Adaptive DNN inference with early exiting leverages the observation that some test examples can be easier to predict than others. This paper presents EENet, a novel early-exiting scheduling framework for multi-exit DNN models. Instead of having every sample go through all DNN layers during prediction, EENet learns an early exit scheduler, which can intelligently terminate the inference earlier for certain predictions, which the model has high confidence of early exit. As opposed to previous early-exiting solutions with heuristics-based methods, our EENet framework optimizes an early-exiting policy to maximize model accuracy while satisfying the given per-sample average inference budget. Extensive experiments are conducted on four computer vision datasets (CIFAR-10, CIFAR-100, ImageNet, Cityscapes) and two NLP datasets (SST-2, AgNews). The results demonstrate that the adaptive inference by EENet can outperform the representative existing early exit techniques. We also perform a detailed visualization analysis of the comparison results to interpret the benefits of EENet.
\end{abstract}

\section{Introduction}

Deep neural networks (DNNs) have shown unprecedented success in various fields such as computer vision and natural language processing, thanks to the advances in computation technologies (GPUs, TPUs) and the increasing amount of available data to train very large and deep neural networks. However, these models usually have very high computational costs, which leads to many practical challenges in deployment and inference on edge computing applications, especially for edge clients with limited resources such as smartphones, IoT devices, and embedded devices~\cite{goodfellow2016deep,adaptive_laskar,ddnoce}. Significant research has been dedicated to improving the computational efficiency of DNN models through training phase optimizations, such as model quantization and deep compression~\cite{quant}, neural network pruning~\cite{prune,lottery}, knowledge distillation~\cite{know_dist}, and multi-exit DNNs~\cite{msdnet,branchynet}. Among these, multi-exit DNNs stand out as a promising technique since they enable adaptive inference with early exiting to reduce inference latency based on the available budget of the deployed device~\cite{adaptive_laskar}.

Early exiting employs the idea of injecting early exit classifiers into certain intermediate layers of a deep learning model and with that multi-exit DNN, gaining the capability to adaptively stop inference at one of these early exits in runtime~\cite{adaptive_laskar}. At the model training phase, we train the DNN model with the additional multiple exit branches through joint loss optimization. During adaptive inference, the multi-exit DNN model can elastically adjust how much time to spend on each sample based on an early exit scheduling policy to maximize the overall performance metric under the given total latency budget. Even though there is a significant line of research on improving the performance of multi-exit DNNs through designing specialized architectures ~\cite{msdnet,supernet,resaware,depthadapt} and training algorithms~\cite{improved_adaptive,distil_exit}, work on optimizing early exit policies is very limited. In the literature, most methods still consider hand-tuned confidence measures for sample scoring such as maximum score~\cite{msdnet}, entropy~\cite{branchynet}, voting~\cite{pabee} etc., and heuristics-based threshold computation approaches.

We argue that through optimizing an early exit scheduling policy, the potential of early exiting can be maximized to efficiently utilize the provided inference resources on heterogeneous edge devices. To this end, we present EENet (Early-Exiting Network), an early exit scheduling framework with two novel functionalities. First, EENet introduces the concept of exit scores by jointly evaluating and combining two complimentary statistics: (i) multiple confidence measures and (ii) class-wise prediction scores. This enables EENet to calibrate output scores for handling statistical differences among prediction scores for different classes. Second, EENet optimizes the distribution of samples to different exits and computes the exit thresholds given the inference budget. We note that the optimization of EENet is performed on the validation dataset in a decoupled manner with the multi-exit DNN training phase. This enables EENet to quickly adapt to varying budget scenarios by only training the schedulers and eliminates the requirement of re-training the full DNN model each time the inference budget changes as opposed to other studies that integrate two optimization phases~\cite{epnet,learnstop}. Lastly, EENet is model-agnostic and applicable to all pre-trained multi-exit models. In addition, flexible deployment is applicable for edge clients with heterogeneous computational resources by running only a partial model split until a certain early exit that matches the resource constraint. 


We conduct extensive experiments with multiple DNN architectures (ResNet~\cite{resnet}, DenseNet~\cite{densenet}, MSDNet~\cite{msdnet}, HRNet~\cite{hrnet}, BERT~\cite{bert}) on three image recognition benchmarks (CIFAR10, CIFAR100, ImageNet), one image segmentation benchmark (Cityscapes) and two sentiment analysis benchmarks (SST-2, AgNews). We demonstrate the improvements of EENet compared to existing representative approaches, such as BranchyNet~\cite{branchynet}, MSDNet~\cite{msdnet}, PABEE~\cite{pabee} and MAML-stop~\cite{learnstop}. Lastly, we provide an ablation study of EENet design components and visual analysis to interpret the behavior of our approach.

\section{Related Work}
\label{sec:rw}

BranchyNet~\cite{branchynet} is the first to explore the idea of early exits for adaptive inference with DNNs. It considers the entropy of prediction scores as the measure of confidence and sets the early exit thresholds heuristically. Certain studies consider maximum prediction score instead of entropy as the exit confidence measure~\cite{msdnet} on computer vision tasks. PABEE~\cite{pabee} proposes stopping the inference when the number of predictions on the same output reaches a certain patience threshold. However, this method depends on having a high number of early exits to produce meaningful scores so that the samples can be separated with a higher resolution for the exit decision. All these methods introduce heuristics-based rules, which do not optimize the early exit policy in terms of scoring and threshold computation. 

Some recent efforts propose task-dependent confidence measures~\cite{mood,dynamicdda} or modifying the training objective to include exit policy learning during the training of the multi-exit DNN~\cite{epnet,learnstop}. EPNet~\cite{epnet} proposes using Markov decision processes however, they add an early exit classifier at each exit to increase the number of states, which is computationally unfeasible since each early exit introduces an additional computational cost during both training and inference. MAML-stop~\cite{learnstop} proposes a variational Bayesian approach to learning when to stop predicting during training. The major drawback of these approaches is the requirement of training the full model for every different budget value since the optimization of early exiting behavior is integrated into the DNN training phase.

\section{Methodology}
\label{sec:method}

\begin{figure*}[t!]
    \centering
    \includegraphics[width=\textwidth]{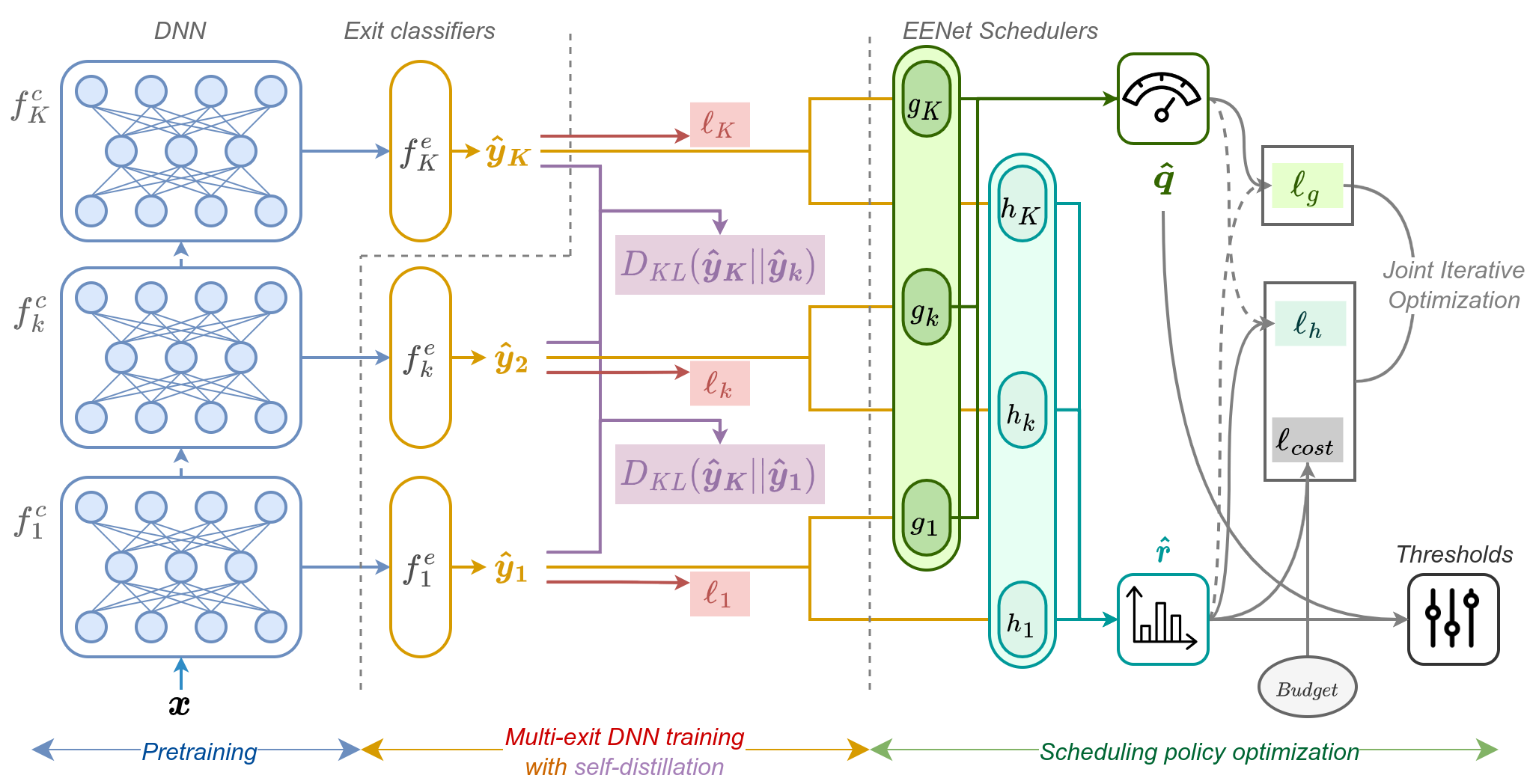}
    \caption{\small Architectural overview of EENet. The training phase contains the steps necessary to obtain and train the multi-exit DNN model. The scheduling optimization phase involves optimizing the early exit schedulers and computing exit thresholds given the inference budget. Each scheduler at exit $k$ encapsulates an exit scoring function $g_k$, which outputs an exit score $\hat{q}_k$ representing the confidence in the correctness of the model prediction, and an exit assignment function $h_k$, which is used to estimate the posterior distribution over the exit assignment $\boldsymbol{\hat{r}}$.}
    \label{fig:arch}
\end{figure*}

\paragraph{Overview.} Figure~\ref{fig:arch} provides an architectural overview of our system. Given a pre-trained DNN model, we first inject early exits and finetune the multi-exit model on the training dataset. During this training phase, we jointly optimize the losses from each exit. During multi-exit model training to enhance the performance of earlier exits, we perform self-distillation between early exits and the final exit by minimizing the KL divergence between outputs. Next, we initiate the early exit scheduling policy optimization phase on the validation dataset. The goal of this step is to optimize the exit scoring functions and exit thresholds such that the provided inference budget is satisfied and the performance metric is maximized. We first obtain the validation predictions of the trained multi-exit model and construct the training dataset for scheduling optimization. We iteratively optimize the exit scoring functions ($g_k$) and exit assignment functions ($h_k$). Finally, we compute the early exit triggering thresholds, one for each exit, and complete the early exit scheduling policy optimization phase. The trained multi-exit model, with the exit scoring functions and exit thresholds, is then utilized during the test to determine which exit each sample should take. In the rest of this section, we provide the multi-exit model training in Section~\ref{sec:train}. Then, we describe the adaptive inference with early exiting and scheduling optimization methodology in Section~\ref{sec:opt}.

\subsection{Multi-exit Model Training with Self-distillation}
\label{sec:train}
To enable a given pre-trained DNN classifier to perform early exiting, we first inject early exit classifiers into the model at certain intermediate layers. If the configuration of per-client computational resource constraints, which specifies the computational budget $B_m$ of a client $m \in \{1, ..., M\}$, is available, the number of exits $K$ can be determined heuristically with clustering analysis of the computational requirements of potential users $\{B_m | 1\leq m\leq M\}$. In this work, we set $K$ based on the number of resource categories of heterogeneous edge clients. Then, we set the exit locations $l_k$ for each exit $k$ based on the lowest resource capacity from the group of edge clients in the $k$th resource category by finding the largest submodel $f_k$ while not exceeding the constraints satisfying that $\texttt{cost}(f_k) \leq \min \{B_m | m \in \mathcal{S}_k\}$, where $\mathcal{S}_k$ is the set of clients at the $k$th resource category and $\texttt{cost}$ is the cost function (\#PARAMS, \#FLOPs, latency etc.).

Let $f$ denote a multi-exit classification model capable of outputting multiple predictions after the injection of early exit subnetworks $f^{e}_k$ at each exit $k$. 
After preparing the multi-exit model with $K$ exits, we start finetuning with joint loss optimization across all $K$ exit classifiers. Let us denote the set of output probability scores of $f$ for one sample as $\{\boldsymbol{\hat{y}}_k\}_{k=1}^K$ and the label as $y \in \mathcal{C}$, where $K$ is the number of exits and $\mathcal{C} = \{1, 2, \dots C\}$ is the set of classes. Here, at each exit $k$, $\boldsymbol{\hat{y}}_k = f_k(\boldsymbol{x}) = [\dots \hat{y}_{k,c} \dots] \in \mathbb{R}^C$ is the vector of prediction scores for each class $c$, where $f_k \triangleq f^{e}_k \circ f^{c}_k \circ \hdots f^{c}_1$ with $f^{c}_k$ the $k$th core subnetwork and $f^{e}_k$ the $k$th early exit subnetwork of the multi-exit model $f$.

During the training/finetuning of these $K$-exit models, we minimize the weighted average of cross-entropy losses from each exit combined with the KL divergence among exit predictions for self-distillation: 
$
    \mathcal{L}_{train} = \sum_{k=1}^K \gamma_k \ell_k +\alpha_{KL}\sum_{k=1}^{K-1} D_{KL}(\boldsymbol{\hat{y}}_K || \boldsymbol{\hat{y}}_k),
$
where $\ell_k$ is the cross-entropy loss at the $k$th exit, $\gamma_k= \frac{k}{K(K+1)}$ is the loss weight of the $k$th exit, and $D_{KL}(\boldsymbol{\hat{y}}_K || \boldsymbol{\hat{y}}_k) = \texttt{sum}(\sigma(\boldsymbol{\hat{y}}_K / \tau) \log{\frac{\sigma(\boldsymbol{\hat{y}}_{K} / \tau)}{\sigma(\boldsymbol{\hat{y}}_{k} / \tau)}})\tau^2$ is the forward KL divergence from $\boldsymbol{\hat{y}}_k$ to $\boldsymbol{\hat{y}}_K$. Here, $\alpha_{KL}$ and $\tau$ are the hyperparameters that control the effect of self-distillation.

\subsection{Adaptive Early-Exit Inference Optimization}
\label{sec:opt}
After training the multi-exit classification model $f$ with $K$ exits, we store the validation predictions of the multi-exit model at each exit to use during scheduler optimization. Then, we move forward to generating an early exit policy under given budget constraints. 
\paragraph{Problem Definition.} We are given an average per-sample inference budget $B$ (in terms of latency, \#FLOPs etc.), and the vector of computation costs $\boldsymbol{c} \in \mathbb{R}^K$ of the model $f$ until each exit. On a dataset with $N$ examples, $\mathcal{D} = \{(\{\boldsymbol{\hat{y}}_{n, k}\}_{k=1}^K, y_n)\}_{n=1}^N$ containing model prediction scores and labels on the validation set, the ultimate goal is to find the exit scoring functions ($\{g_k\}_{k=1}^K$) and the thresholds $\boldsymbol{t} \in \mathbb{R}^K$ that maximizes the accuracy such that:
\begin{equation}
        \boldsymbol{t}, \{g_k\}_{k=1}^K = \argmax_{\boldsymbol{t} \in \mathbb{R}^K, \{g_k: \mathbb{R}^D \rightarrow \mathbb{R}\}_{k=1}^K} \frac{1}{N} \sum_{n=1}^N \mathbf{1}_{\hat{y}_{n, k_n} = y_n}
\end{equation}
while satisfying the given average per-sample inference budget $B$ such that $\frac{1}{N} \sum_{n=1}^N c_{k_n} \leq B$. 

\begin{figure}[t!]
    \centering
    \includegraphics[width=\columnwidth]{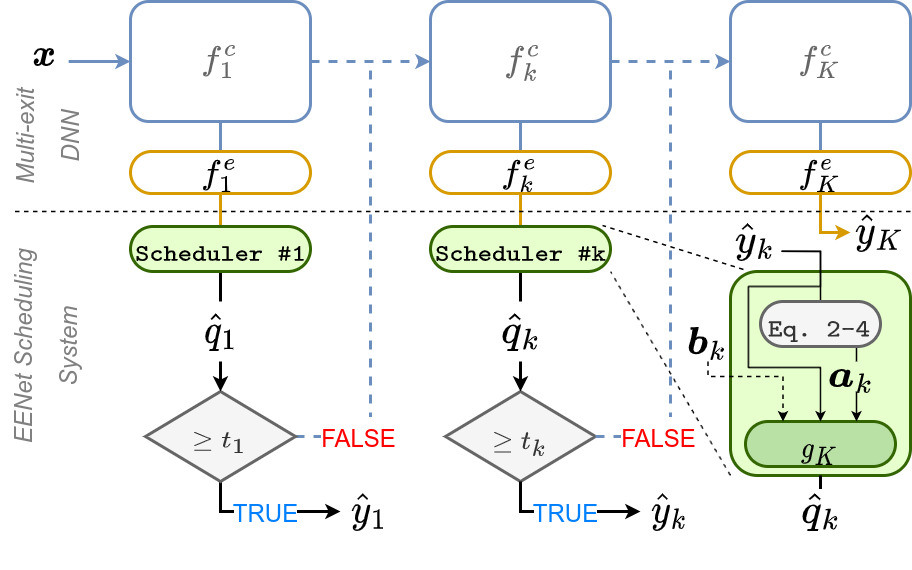}
    \caption{\small Adaptive inference with early exit scheduling. Early exiting at exit $k$ is performed if the computed exit score $\hat{q}_k$ (estimated probability of a correct prediction) is higher than the threshold $t_k$. Execution continues until the condition is met.}
    \label{fig:infer}
    \vspace{-5pt}
\end{figure}

Here, $k_n = min \{k | g_k(\boldsymbol{\hat{y}}_{n, k}) \geq t_k\}$ denotes the minimum exit index where the computed exit score is greater or equal to the threshold of that exit, i.e. the assigned exit for the $n$th sample. Exit scoring functions $g_k$ take the corresponding prediction score vector at exit-$k$ as input and return the exit score for that sample, which represents the likelihood of a correct prediction, i.e. $g_k(\boldsymbol{\hat{y}}) \triangleq P(y = \hat{y} | \boldsymbol{\hat{y}}, k)$. The pair of exit scoring functions ($g_1, g_2 \hdots g_K$) and thresholds ($t_1, t_2, \hdots t_K$) that maximize the validation accuracy while satisfying the given average budget are then used for early-exit enabled inference as will be explained at Section~\ref{sec:infer_opt} and illustrated in Figure~\ref{fig:infer}. 

\subsubsection{EENet Scheduler Architecture} 
\label{sec:infer_opt}
We develop a multi-objective optimization approach with three goals: (i) estimating whether a given prediction is correct or not, (ii) estimating the exit assignment for a given sample, and (iii) satisfying the given average per sample inference budget over the validation dataset. To this end, first, we define a target variable $q_k = \mathbf{1}_{\hat{y}_k=y}$, representing the correctness of a prediction, where $\hat{y}_k \triangleq \argmax_{c \in \mathcal{C}}\hat{y}_{k,c}$ is the predicted class. Here, $q_k$ is equal to one if the prediction at exit $k$ is correct and zero otherwise. In addition, we define a confidence score vector $\boldsymbol{a}_k$ containing three different confidence measures: (i) maximum prediction score, which measures the confidence based on the highest probability score among predictions for each class (ii) entropy, which measures the confidence based on the entropy of probability scores (i.e., if entropy is high, there is high uncertainty therefore, confidence is low) and (iii) voting, which measures the confidence based on the number of agreements among different exit classifiers. These values can be formulated as below:
\begin{align}
    a_{k}^{(max)} &= \max_{c \in \mathcal{C}}\hat{y}_{k,c}, \\
    a_{k}^{(entropy)} &= 1 + \frac{\sum_{c'=1}^C{\hat{y}_{k, c'}\log{\hat{y}_{k, c'}}}}{\log{C}}, \\
    a_{k}^{(vote)} &= \frac{1}{k}\max_{c \in \mathcal{C}}{\sum_{k'=1}^k \mathbf{1}_{\hat{y}_{k'}=c}}. \label{eq_a}
\end{align}

\begin{algorithm}[b!]
\algsetup{linenosize=\normalsize}
\caption{\texttt{EENet Scheduling Optimization}}
    \begin{algorithmic}[1]
        \STATE \textbf{Inputs:} $\mathcal{D} = \{(\{\boldsymbol{\hat{y}}_{n, k}\}_{k=1}^K, y_n)\}_{n=1}^N$, $B$, $\boldsymbol{c} \in \mathbb{R}^K$
    	\STATE \textbf{Outputs:} $\{g_k\}_{k=1}^K$, $\boldsymbol{t} \in \mathbb{R}^{K}$
        \STATE Initialize: $\boldsymbol{h} \leftarrow \texttt{zeros}(N)$, $\boldsymbol{t} \leftarrow \texttt{ones}(k)$
        \FOR{\textit{iter} $i=1$ \TO $N_{iter}$}
            \STATE Optimize $\{g_k\}_{k=1}^K$ by minimizing $\mathcal{L}_g$ on $\mathcal{D}$.
            \STATE Optimize $\{h_k\}_{k=1}^K$ by minimizing $\mathcal{L}_h$ on $\mathcal{D}$.
        \ENDFOR
        \STATE Compute exit scores using $\{g_k\}_{k=1}^K$: $\mathbf{\hat{Q}} \triangleq (\hat{q}_{n, k}) \in \mathbb{R}^{N \times K}$
        \STATE $\mathbf{S} = (s_{n, k}) \in \mathbb{N}^{N \times K} \leftarrow \texttt{argsort}(\mathbf{\hat{Q}}, 1)$
        \FOR{\textit{exit index} $k=1$ \TO $K-1$}
            \STATE $c \leftarrow 0$
            \STATE Estimate exit distribution using $\{h_k\}_{k=1}^K$: $p_k \leftarrow \frac{1}{N}\sum_{n=1}^{N} \hat{r}_{n, k}$
            \FOR{\textit{sample index} $n=1$ \TO $N$} 
                \IF{$h_{s_{n, k}} = 0$}
                    \STATE $c \leftarrow c+1$
                    \STATE $h_{s_{n, k}} \leftarrow 1$ 
                    \IF{$c = \texttt{round}(N p_k)$}
                        \STATE $t_k \leftarrow \hat{q}_{s_{n, k}, k}$
                        \STATE \textbf{break}
                    \ENDIF
                \ENDIF
            \ENDFOR
        \ENDFOR
        \STATE $t_K \leftarrow 0$ 
        \RETURN $\{g_k\}_{k=1}^K$, $\boldsymbol{t}$
	\end{algorithmic} \label{alg_1}
\end{algorithm}

At each exit $k$, using the prediction score vector $\boldsymbol{\hat{y}}_k$ and the confidence score vector $\boldsymbol{a}_k$, EENet calibrates the predictions and computes the exit score $\hat{q}_k$ as a linear combination to estimate the correctness of the prediction: $
    \hat{q}_k = g_k(\boldsymbol{\hat{y}}_k, \boldsymbol{a_k}, \boldsymbol{b}_k) = \texttt{clamp}(\boldsymbol{\psi}^T [\boldsymbol{\hat{y}}_k, \boldsymbol{a_k}, \boldsymbol{b}_k], 0, 1) \triangleq P(y = \hat{y} | \boldsymbol{\hat{y}}, k) \label{eq_q}
$,
where $\boldsymbol{b}_k = [\hat{q}_1, \dots \hat{q}_{k-1}]$ for $k>1$ is the collection of previous exit scores. 

Next, we consider the assigned exit $k$ as latent variable and define the exit assignment functions $h_k$ to estimate the posterior distribution over exits $\hat{r}_k \triangleq p(k|\boldsymbol{\hat{y}})$ as follows:
\begin{equation}
    \Tilde{r}_k = h_k(\boldsymbol{\hat{y}}_k, \boldsymbol{a}_k, \boldsymbol{b}_k), \quad \hat{r}_k = \frac{e^{\Tilde{r}_k}}{\sum_{k'=1}^K e^{\Tilde{r}_{k'}}}, \label{eq_r}
\end{equation}
where $h_k$ is a two-layer MLP with ReLU activation and $D_h$ hidden dimensions. Outputs over all exits are normalized using the softmax function. 

\subsubsection{Optimization}
We provide the pseudocode for the optimization of the exit scoring functions $\{g_k\}_{k=1}^K$ and thresholds $\boldsymbol{t}$ in Algorithm \ref{alg_1}. Here in line 4-6, given the dataset containing model predictions and target labels on the validation data, budget and inference costs, we iteratively optimize the exit scoring and assignment functions by minimizing the losses observed by these functions. 

For the optimization objective of exit scoring functions, we define the loss $\mathcal{L}_g$ as follows:
\begin{equation}
     \mathcal{L}_g = \frac{1}{K}\sum_{n=1}^{N}\sum_{k=1}^K w_{n, k} \ell_g(\hat{q}_{n, k}, q_{n, k}), \label{eq:L_g}
\end{equation}
where $\ell_g(\hat{q}_{k}, q_{k}) = q_k\log(\hat{q}_k) + (1-q_k)\log(1-\hat{q}_k)$ is the binary CE loss. $N$ is the number of validation data samples and $w_{n, k} = \frac{\hat{r}_{n, k}}{\sum_{n'=1}^{N} \hat{r}_{n', k}}$ is the loss weight for the $n$th sample at the $k$th exit. This weighting scheme encourages exit scoring functions to specialize in their respective subset of data. For instance, for a given sample, if the exit assignment score at a particular exit is high, the exit scoring function at that exit will be optimized with that sample more compared to other samples.

For the optimization objective of exit assignment functions, we ideally would try to achieve a distribution that maximizes the following objective:
\begin{equation}
    \argmax_{p(k|\cdot)} \ \mathbb{E}_{k \sim p(k|\cdot)} \log P(y=\hat{y}|\boldsymbol{\hat{y}}, k) + \beta_h H(p). \label{eq_exp}
\end{equation}
In other words, the expected log-probability of a correct prediction should be maximized in addition to an entropy regularization term, $H(p) \triangleq - \sum_k p(k) \log p(k)$, whose effect is controlled with $\beta_h > 0$. Taking the derivative with respect to $p$ yields the following target distribution:
\begin{equation}
    p^*(k|y, \boldsymbol{\hat{y}}_k) = \frac{P(y=\hat{y}|\boldsymbol{\hat{y}}_k, k) ^ {1 / \beta_h}}{\sum_{k'} P(y=\hat{y}|\boldsymbol{\hat{y}}_{k'}, k') ^ {1 / \beta_h}} = 
    \frac{\hat{q}_k ^ {\ 1 / \beta_h}}{\sum_{k'} \hat{q}_{k'} ^ {\ 1 / \beta_h}} \label{eq_p}
\end{equation}

Our goal is to optimize a distribution $p(k | \boldsymbol{\hat{y}}_k)$ that does not require observing the label and it will be as close as possible to this target distribution $p^*(k|y, \boldsymbol{\hat{y}}_k)$ while satisfying the budget requirement. To this end, we define the exit assignment loss as a combination of KL-divergence between $p$ and $p^*$, and a budget loss such that $\mathcal{L}_h = D_{KL}(p^*||p) + \alpha_{cost}\ell_{cost}$, where
\begin{equation}
\label{eq_losskl}
    D_{KL}(p^*||p) = \frac{1}{NK}\sum_{n=1}^N \sum_{k=1}^K p^*(k|y_n, \boldsymbol{\hat{y}}_{n, k}) \log \frac{p^*(k|y_n, \boldsymbol{\hat{y}}_{n, k})}{p(k | \boldsymbol{\hat{y}}_{n, k})},  \\  \label{eq_lossce}
    \ell_{cost} = \frac{1}{B} | B - \frac{1}{N}\sum_{n=1}^{N}\sum_{k=1}^{K}\hat{r}_{n, k}c_k |.
\end{equation}
$p(k | \boldsymbol{\hat{y}}_{n, k}) = \hat{r}_{n, k}$ is provided in Eq.~\ref{eq_r} and $p^*(k|y_n, \boldsymbol{\hat{y}}_{n, k})$ is provided in Eq.~\ref{eq_p}. Here, $\ell_{cost}$ is the budget cost and calculates the scaled absolute distance between the inference budget and used resources using the computed exit assignment scores and inference costs until each exit. 

After optimizing the exit scoring functions $\{g_k\}_{k=1}^K$ and exit assignment functions $\{h_k\}_{k=1}^K$, we compute the exit thresholds. We first sort the exit scores for all samples at each exit (line 8). Then at each exit, we let $Np_k$ samples with the highest scores exit, where $p_k = \frac{1}{N}\sum_{n=1}^{N} \hat{r}_{n, k}$ is the mean of exit assignment scores of validation samples. And, we set the threshold to the exit score of the last exited sample with the lowest score (Algorithm~\ref{alg_1}, lines 9-19).

We provide the formulation of our framework for the traditional classification tasks. For dense prediction tasks such as image segmentation, we consider the mean of the pixel-level scores during exit score and assignment computations. Regression tasks such as object detection/tracking (early exiting for easier frames with fewer objects/less occlusion, etc.) are also applicable with minor modifications: (i) instead of $\boldsymbol{\hat{y}}_k$, we can use the hidden layer output $(f^{c}_k \circ \hdots f^{c}_1)(\boldsymbol{x})$ as input to $g_k$ and $h_k$, (ii) we can treat exit scoring functions as boosters by defining residuals as their targets, and (iii) for MSE loss, we can consider Gaussian distribution to derive the likelihood for Eq.~\ref{eq_exp}.

\begin{table*}[t]
\centering
\setlength\tabcolsep{4pt}
\resizebox{\textwidth}{!}{%
{\renewcommand{\arraystretch}{1.05}%
\begin{tabular}{l|c|c|c|ccccc}
\multirow{2}{*}{\begin{tabular}[c]{@{}l@{}}Dataset, Model and \\ Base Performance \end{tabular}} & \multicolumn{1}{c|}{\multirow{2}{*}{\begin{tabular}[c]{@{}c@{}}Base Model \\ Latency \end{tabular}}} & \multicolumn{1}{c|}{\multirow{2}{*}{\begin{tabular}[c]{@{}c@{}}Average Latency \\ Budget per sample \end{tabular}}} & \multicolumn{1}{c|}{\multirow{2}{*}{\begin{tabular}[c]{@{}c@{}}Speed Gain \end{tabular}}} & \multicolumn{5}{c}{Metric (\%)}\\ \cline{5-9}
& \multicolumn{1}{c|}{}& \multicolumn{1}{c|}{}& \multicolumn{1}{c|}{}& \textbf{BranchyNet} & \textbf{MSDNet} & \textbf{PABEE} & \textbf{MAML-stop} & \textbf{EENet} \\ \hline
\multirow{3}{*}{\begin{tabular}[c]{@{}l@{}}\textbf{CIFAR-10}\\ ResNet56 w/ 3 exits\\ 93.90\% \textit{- accuracy}\end{tabular}} & \multirow{3}{*}{\begin{tabular}[c]{@{}l@{}}4.70 ms\end{tabular}} & 3.50 ms & 1.34x & 93.76 & 93.81 & 93.69& - &\textbf{93.84} $\pm$ \text{0.05}\\
& & 3.00 ms & 1.56x & 92.57 & 92.79 & 91.85& - &\textbf{92.90} $\pm$ \text{0.13}\\
& & 2.50 ms & 1.88x & 87.55 & 88.76 & 84.39& 88.67&\textbf{88.90} $\pm$ \text{0.19}\\ \hline
\multirow{3}{*}{\begin{tabular}[c]{@{}l@{}}\textbf{CIFAR-100}\\ DenseNet121 w/ 4 exits\\ 75.08\% \textit{- accuracy}\end{tabular}} & \multirow{3}{*}{\begin{tabular}[c]{@{}l@{}}10.20 ms\end{tabular}} & 7.50 ms & 1.36x & 73.96 & 74.01 & 73.68& -& \textbf{74.08} $\pm$ \text{0.13}\\
& & 6.75 ms & 1.51x & 71.65 & 71.99 & 68.10& -&\textbf{72.12} $\pm$ \text{0.20}\\
& & 6.00 ms & 1.70x & 68.13 & 68.70 & 61.13& 69.00&\textbf{69.57} $\pm$ \text{0.22}\\ \hline
\multirow{3}{*}{\begin{tabular}[c]{@{}l@{}}\textbf{ImageNet}\\ MSDNet35 w/ 5 exits\\ 74.60\% \textit{- accuracy}\end{tabular}} & \multirow{3}{*}{\begin{tabular}[c]{@{}l@{}}195.14 ms\end{tabular}} & 125.00 ms & 1.56x & 74.10 & 74.13 & 74.05& -&\textbf{74.18} $\pm$ \text{0.13} \\
& & 100.00 ms & 1.95x & 72.44 & 72.70 & 72.40& -&\textbf{72.75} $\pm$ \text{0.11} \\
& & 75.00 ms & 2.60x & 69.32 & 69.76 & 68.13& 69.55&\textbf{69.88} $\pm$ \text{0.15}\\ \hline
\multirow{3}{*}{\begin{tabular}[c]{@{}l@{}}\textbf{Cityscapes}\\ HRNET-W48 w/ 3 exits\\ 80.90\% \textit{- mIoU}\end{tabular}} & \multirow{3}{*}{\begin{tabular}[c]{@{}l@{}}131.60 ms\end{tabular}} & 100.00 ms & 1.32x & 79.22 & 78.75 & 72.20 & -& \textbf{80.03} $\pm$ \text{0.21}\\
& & 50.00 ms & 2.63x & 75.54 & 75.50 & 70.09 & - &\textbf{76.90} $\pm$ \text{0.17}\\
& & 15.00 ms & 8.77x & 68.12 & 68.35 & 63.95 & 65.76 &\textbf{70.30} $\pm$ \text{0.35}
\end{tabular}}}
\caption{\small Image classification/segmentation experiment results in terms of accuracy for image classification on CIFAR-10, CIFAR-100, ImageNet and mean IoU for image segmentation on Cityscapes dataset at different average latency budget values.}\label{tab:results_cv}
\end{table*}
\begin{table*}[t!]
\centering
\setlength\tabcolsep{4pt}
\resizebox{\textwidth}{!}{%
{\renewcommand{\arraystretch}{1.05}%
\begin{tabular}{l|c|c|c|ccccc}
\multirow{2}{*}{\begin{tabular}[c]{@{}l@{}}Dataset, Model and \\ Base Performance \end{tabular}} & \multicolumn{1}{c|}{\multirow{2}{*}{\begin{tabular}[c]{@{}c@{}} Base Model \\ Latency \end{tabular}}} & \multicolumn{1}{c|}{\multirow{2}{*}{\begin{tabular}[c]{@{}c@{}}Average Latency \\ Budget per sample \end{tabular}}} & \multicolumn{1}{c|}{\multirow{2}{*}{\begin{tabular}[c]{@{}c@{}}Speed Gain \end{tabular}}} & \multicolumn{5}{c}{Metric (\%)}\\ \cline{5-9} 
& \multicolumn{1}{c|}{}& \multicolumn{1}{c|}{}& \multicolumn{1}{c|}{}& \textbf{BranchyNet} & \textbf{MSDNet} & \textbf{PABEE} & \textbf{MAML-stop} & \textbf{EENet} \\ \hline
\multirow{3}{*}{\begin{tabular}[c]{@{}l@{}}\textbf{SST-2}\\ BERT-base w/ 4 exits\\ 92.36\% \textit{- accuracy} \end{tabular}}& \multirow{3}{*}{\begin{tabular}[c]
{@{}l@{}}189.93 ms\end{tabular}} & 150.00 ms & 1.27x & 92.14 & 92.17 & 92.05&  -& \textbf{92.25} $\pm$ \text{0.03} \\
& & 125.00 ms & 1.52x & 90.86 & 91.00 & 90.75& -&\textbf{92.09} $\pm$ \text{0.05} \\
& & 100.00 ms & 1.89x & 87.66 & 87.71 & 86.99& 88.15&\textbf{91.58} $\pm$ \text{0.60}  \\ \hline
\multirow{3}{*}{\begin{tabular}[c]{@{}l@{}}\textbf{AgNews}\\ BERT-base w/ 4 exits\\ 93.98\% \textit{- accuracy} \end{tabular}}& \multirow{3}{*}{\begin{tabular}[c]{@{}l@{}}189.93 ms\end{tabular}} & 150.00 ms & 1.27x & 93.20 & 93.17 & 93.15 & - &\textbf{93.85} $\pm$ \text{0.01} \\
& & 125.00 ms & 1.52x & 92.95 & 92.98 & 92.57& -& \textbf{93.75} $\pm$ \text{0.11} \\
& & 100.00 ms & 1.89x & 85.58 & 84.93 & 85.22& 93.00& \textbf{93.45} $\pm$ \text{0.09} 
\end{tabular}}}
\caption{\small Sentiment analysis experiment results in terms of accuracy obtained at different average latency budget values.} \label{tab:results_nlp}
\end{table*}

\section{Experiments}
\label{sec:exp}

We conduct experiments with convolutional and transformer-based networks to evaluate EENet and report the performance improvements obtained for budget-constrained adaptive inference on six image/text benchmarks (CIFAR-10, CIFAR-100, ImageNet, Cityscapes, SST-2 and AgNews). We provide detailed explanations of experiment setups, and further analysis in the supplementary material. Our code is available at \href{https://github.com/git-disl/EENet}{https://github.com/git-disl/EENet}.

\subsection{Datasets and Preprocessing} \noindent In image classification experiments, we work on CIFAR-10/100~\cite{cifar} and ImageNET~\cite{imagenet} datasets. CIFAR-10 and CIFAR-100 contain 50000 train and 10000 test images with 32x32 resolution from 10 and 100 classes respectively. ImageNET contains 1.2 million train and 150000 validation images (used for testing) with 224x224 resolution from 1000 classes. We hold out randomly selected 5000 images from CIFAR-10/100 train set and 25000 images from the ImageNET train set for validation. We follow the data augmentation techniques applied in~\cite{resnet}, zero padding, center cropping and random horizontal flip with 0.5 probability. For image segmentation, we use the Cityscapes~\cite{cityscapes} dataset, which contains 5000 (train: 2975, val: 500, test: 1525) images with size 1024x2048, finely labeled in pixel-level for 19 classes. In text classification experiments, we consider SST-2~\cite{sst2} and AGNews~\cite{agnews} datasets. SST-2 contains 67349 train, 872 validation and 1821 samples with positive or negative labels. AGNews contains 120000 train and 7600 samples from four classes. We hold randomly selected 5000 sentences for validation.

\subsection{Validation of EENet with Comparison} We compare our method with various representative early exiting methodologies such as BranchyNet~\cite{branchynet}, PABEE~\cite{pabee}, MSDNet~\cite{msdnet} and MAML-stop~\cite{learnstop} in terms of the accuracy obtained under different latency budget constraints. We consider average latency per sample as the budget definition throughout the experiments. For MSDNet~\cite{msdnet}, BranchyNet~\cite{branchynet} and PABEE~\cite{pabee}, we use maximum score, entropy-based and agreement-based scores as provided in Eq.~\ref{eq_a}. To compute the thresholds for these methods, we follow the approach in \cite{msdnet} and assume that the exit assignment of samples will follow a geometric distribution since BranchyNet and PABEE do not provide any guidance on how to set the exit thresholds.

We provide Table~\ref{tab:results_cv} for the results on image classification and Table~\ref{tab:results_nlp} for sentiment analysis tasks. Since MAML-stop requires training the full DNN model for each different budget setting, we obtain the result for one budget value in each experiment. Please note that we refer to the maximum prediction score-based early exiting approach in \cite{msdnet} as MSDNet and the multi-exit CNN architecture with 35 layers as MSDNet35. As demonstrated in Table~\ref{tab:results_cv}, our system consistently performs better compared to other early exit approaches for image tasks. In addition, we observe that our approach achieves greater performance gains as the budget tightens with improvements ranging from 0.23\% to 1.95\% compared to the closest competitor method. We observe that MAML-stop performs closest in most scenarios but requires finetuning the full multi-exit model separately for each budget setting, which is not practical in most applications. PABEE only considers the agreement among different early exit classifiers and completely ignores the scores, which causes it to perform worse in most settings. 
Similar observations are also made for the sentiment analysis task, showing that EENet offers consistent performance improvement for all six benchmarks across different budget settings thanks to optimizing exit scoring and distribution as opposed to manual rule-based approaches. For instance, our approach achieves 93.45\% accuracy on AgNews at 100ms/sample while BranchyNet can only achieve 93.25\% even at 150ms/sample.

\begin{figure*}[t!]
    \centering
    \includegraphics[width=0.95\textwidth]{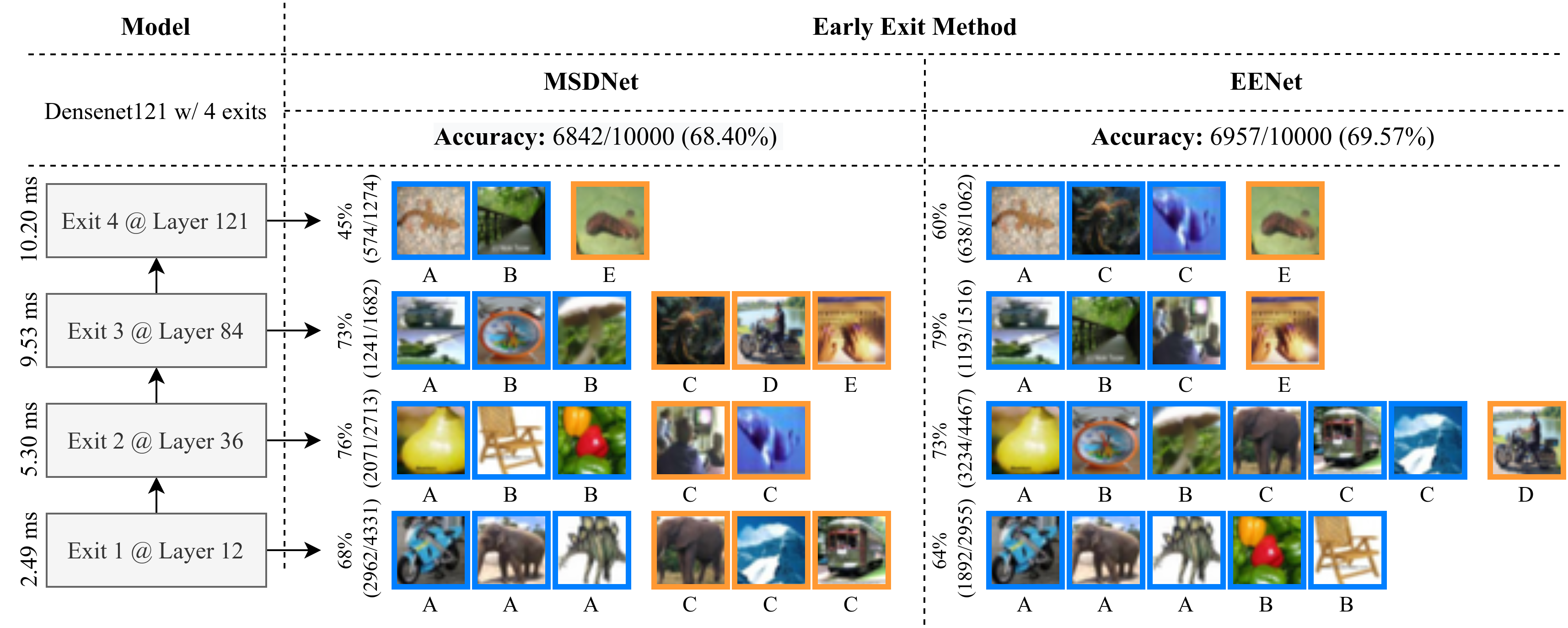}
    \caption{\small Visual comparison of MSDNet and EENet results on CIFAR-100 for the average latency budget of 6 ms. We illustrate randomly selected twenty samples and the exit location that they were assigned. Images with blue/orange borders are predicted correctly/incorrectly at the corresponding exit. We also report the number of correct predictions and exited samples at each exit.}
    \label{fig:comp1}
\end{figure*}

\begin{figure*}[t!]
    \centering
    \includegraphics[width=0.95\textwidth]{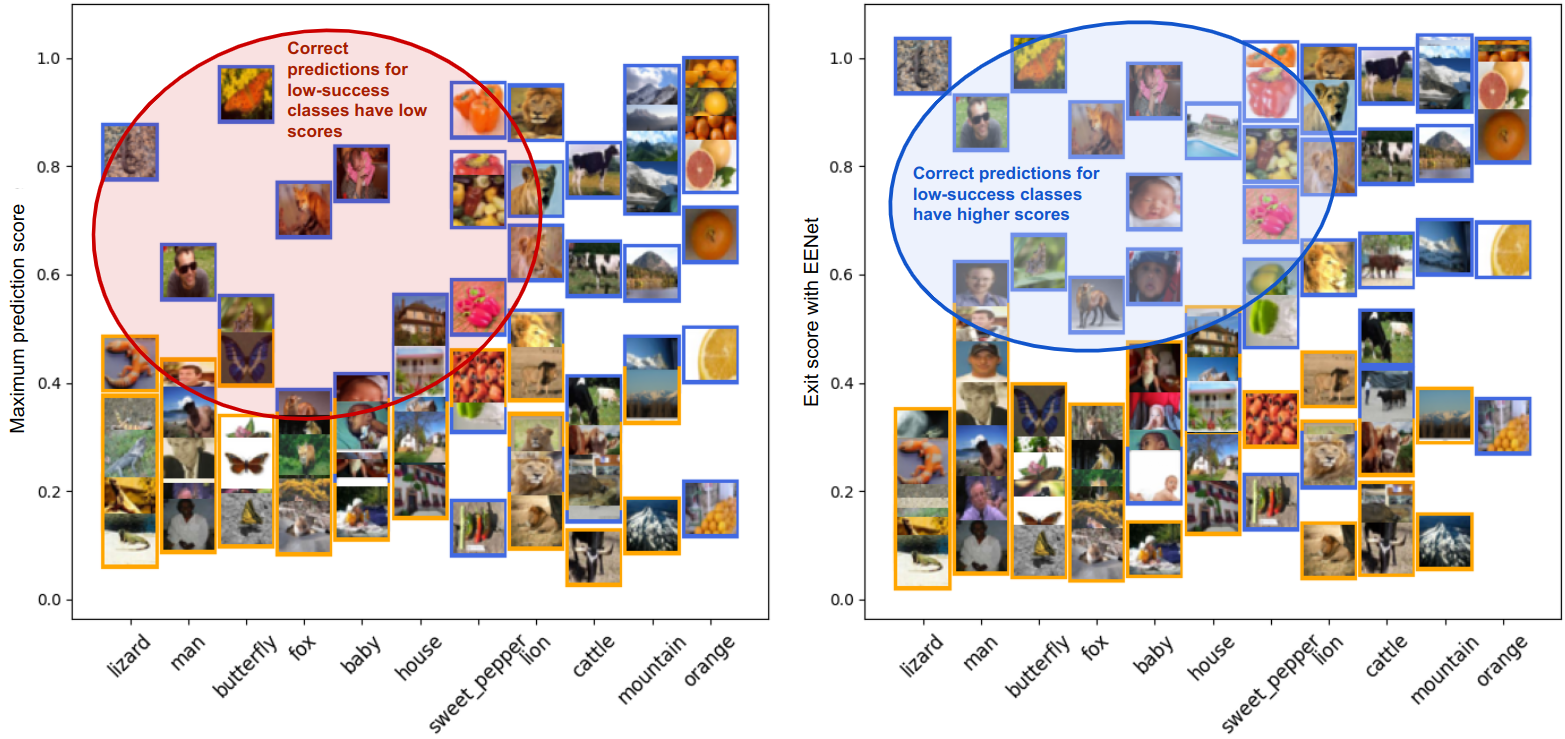}
    \caption{\small Images with blue/orange borders are predicted correctly/incorrectly at the first exit of DenseNet121 on CIFAR-100 (left: MSDNet, right: EENet). Our approach provides a clearer separation of true and false predictions for all classes, compared to maximum prediction score-based confidence, which is popularly used in the literature. Correct predictions for low-success classes have lower maximum prediction scores whereas their exit scores computed by EENet are higher as desired.}
    \label{fig:visual}
\end{figure*}

\begin{table*}[]
\centering
\setlength\tabcolsep{1.5pt}
\resizebox{\textwidth}{!}{%
{\renewcommand{\arraystretch}{1.25}%
\begin{tabular}{l|cc|cc|cc|cc|cc|cc}
\multirow{2}{*}{Model} & \multicolumn{2}{c|}{\textbf{Exit-1}} & \multicolumn{2}{c|}{\textbf{Exit-2}} & \multicolumn{2}{c|}{\textbf{Exit-3}} & \multicolumn{2}{c|}{\textbf{Exit-4}} & \multicolumn{2}{c|}{\textbf{Exit-5}} & \multicolumn{2}{c}{\textbf{Base Model}} \\
 & \#PRMs& Latency& \#PRMs& Latency& \#PRMs& Latency& \#PRMs& Latency& \#PRMs& Latency& \#PRMs & Latency\\ \hline
\multirow{2}{*}{\begin{tabular}[c]{@{}l@{}}\textbf{ResNet56}\\ (w/ EENet)\end{tabular}}& 0.06M & 2.31ms& 0.28M & 4.15ms & 0.96M & 4.93ms & - & -& - & -& 0.86M& 4.77ms \\
 & (\textless{}1K)& (+0.01ms) & (\textless{}1K)& (+0.01ms) & (\textless{}1K)& (+0.01ms) & - & -& - & -& -& -\\ \hline
\multirow{2}{*}{\begin{tabular}[c]{@{}l@{}}\textbf{DenseNet121}\\ (w/ EENet)\end{tabular}} & 0.06M & 2.49ms& 0.25M & 5.30ms& 0.86M & 9.53ms & 1.17M & 10.20ms & - & -& 1.04M& 10.03ms \\
 & (+5.25K)& (+0.08ms) & (+5.36K)& (0.08ms) & (+5.47K)& (+0.08ms) & (+5.57K)& (+0.08ms) & - & -& -& -\\ \hline
\multirow{2}{*}{\begin{tabular}[c]{@{}l@{}}\textbf{MSDNet35}\\ (w/ EENet)\end{tabular}}& 8.76M & 58.95 ms& 20.15M& 122.99 ms& 31.73M& 155. 49 ms& 41.86M& 177.69 ms& 62.31M& 194.31 ms& 58.70M & 188.49 ms\\
 & (+0.25M)& (+0.72ms) & (+0.25M)& (+0.72ms) & (+0.25M)& (+0.72ms) & (+0.25M)& (+0.72ms) & (+0.25M)& (+0.72ms) & -& -\\ \hline
\multirow{2}{*}{\begin{tabular}[c]{@{}l@{}}\textbf{HRNet-W48}\\ (w/ EENet)\end{tabular}}& 1.59M & 4.80 ms& 17.14M& 19.17 ms& 65.96M& 138.05 ms& - & - & - & - & 63.60M & 131.60 ms\\
 & (\textless{}1K)& (+1.21ms) & (\textless{}1K)& (+1.21ms) & (\textless{}1K)& (+1.24ms) & - & - & - & - & -& -\\ \hline
\multirow{2}{*}{\begin{tabular}[c]{@{}l@{}}\textbf{BERT-base}\\ (w/ EENet)\end{tabular}}& 45.69M& 51.04 ms& 67.55M& 91.35 ms& 89.40M& 148.13 ms& 111.26M & 188.90 ms& - & -& 109.90M& 183.45 ms\\
 & (\textless{}200)& (\textless{}0.01ms) & (\textless{}200)& (\textless{}0.01ms) & (\textless{}200)& (\textless{}0.01ms) & (\textless{}200)& (\textless{}0.01ms) & - & -& -& - 
\end{tabular}}}
\caption{\small In each row, the top values are the model size (\#PRMs) and average inference latency (ms) measured when running the multi-exit model at inference time. The bottom values in parentheses are the additional cost of model size and latency measured when running the EENet adaptive scheduler to predict the exit scores for each test example. The last column is the model size without early exits.} \label{tab:model_stats}
\end{table*}

\subsection{Analysis of EENet Scheduler Behavior}

\paragraph{Analysis of Exit Assignments.} Figure \ref{fig:comp1} provides a comparison of our method with MSDNet on CIFAR-100, with four exits of the respective early exit models. We randomly display samples and the exit location that they were assigned by our system (right) and by MSDNet (left). Images with green/red borders are predicted correctly/incorrectly at the corresponding exit. Our approach obtains the performance gain over MSDNet by allowing more correct predictions to exit earlier at the second exit. In this figure, Group A is correctly predicted by both at the same exit. Group B samples are correctly predicted by both but EENet exits earlier and gains throughput. Likewise, Group D samples are incorrectly predicted by both but EENet exits earlier. EENet exits later for group C samples to achieve the correct prediction while the other method fails. Group E samples are incorrectly predicted by both at the same exit.

\begin{figure}[]
    \centering
    \includegraphics[width=\columnwidth]{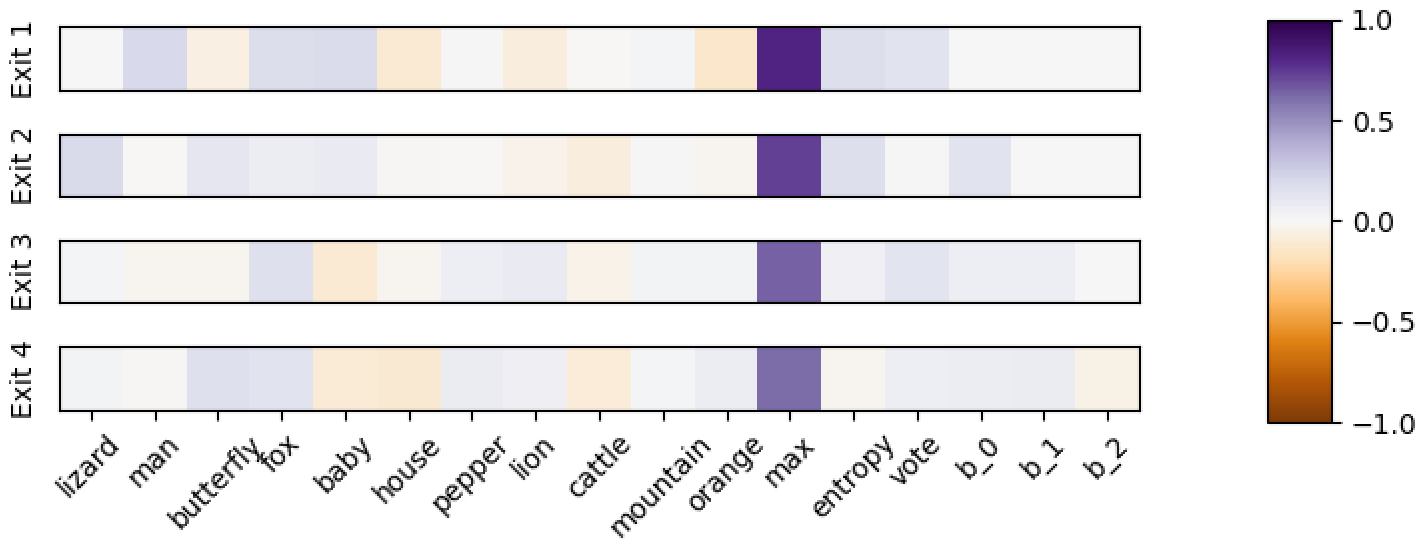}
    \caption{\small Scheduler weights (for randomly selected ten classes and other inputs) of exit scoring functions at each exit at 6.5 ms/sample budget setting for CIFAR-100.}
    \label{fig:inter_weights}
\end{figure}

\paragraph{Analysis of Exit Scores.}
In Figure \ref{fig:visual}, we analyze examples to show why the exit scores produced by our approach are more effective by comparing with the maximum prediction scoring method for early exit used in MSDNet and others in the literature. Randomly selected ten classes are listed on the x-axis sorted by the accuracy achieved using the full model on the corresponding class. From the left figure, using maximum prediction scores to determine exiting may lead to missing some good early exit opportunities. For example, consider those classes that the DNN model produces relatively low maximum prediction scores (less than 0.7), such as lizard, man, butterfly and fox. Even though the predictor can predict them correctly, the relative confidence is not very high. In comparison, the exit scores obtained by our system reflect the correctness of test predictions more accurately. For example, those classes that have lower maximum prediction scores on the true predictions, such as lizard, man, butterfly and fox, will have high exit scores in EENet as shown in the right figure highlighted in the blue oval. To interpret what contributes most during exit score computation, we also analyze the scheduler weights at each exit for CIFAR-100 experiments in Figure~\ref{fig:inter_weights}. The results show that the maximum score contributes the most and higher weights for low-success classes also support the observations.

\subsection{Computational Cost Analysis} \label{sec:edge}
Table \ref{tab:model_stats} reports the number of parameters (\#PRMs) and latency of the models used in the experiments until each exit for four multi-exit DNNs. For each model, we also provide the additional computational cost of EENet in employing the budgeted adaptive early exit policy. The increase of latency caused is negligible ($<0.5\%$) compared to the cost of the forward pass of the original pre-trained DNN model. The scheduler training cost is also very low thanks to lightweight scheduler architectures. For instance, optimizing the schedulers takes less than five minutes in CIFAR experiments on an RTX3060 GPU. 

\paragraph{Partial deployment on edge device hierarchies.} For the edge application scenarios with strict constraints (storage/RAM limitations), the partial model split until a certain exit can be deployed, with the partial model size meeting the edge deployment constraints. With flexible deployment on heterogeneous edge clients, those test samples with exit scores below the learned exit threshold can be passed to the next-level edge server in the hierarchical edge computing infrastructure, which has a higher computational capacity to continue inference.

\section{Conclusion}
We present EENet, an early exit scheduling optimization method for adaptive DNN inference. This paper makes two novel contributions. EENet optimizes both exit scoring functions and exit thresholds such that the given inference budget is satisfied while the performance metric is maximized. As opposed to previous manually defined heuristics-based early exit techniques, including task-specific architectural optimization techniques, our approach is model-agnostic and can easily be used in different deep learning tasks with multi-exit neural networks, ranging from computer vision to NLP applications. In addition, the efficient structure of schedulers provides flexibility compared to the methods that require full model finetuning for varying budget settings. Extensive experiments on six benchmarks  demonstrate that EENet offers significant performance gains compared to existing methods, especially under tighter budget regimes and large DNN models. 

\section*{Acknowledgements}
This research is partially sponsored by a grant from the CISCO Edge AI program, and the NSF CISE grants 2302720, 2312758, and 2038029.

\newpage
{\small
\bibliographystyle{ieee_fullname}
\bibliography{refs}
}

\textbf{}
\newpage
\textbf{}
\newpage
\section*{Supplementary Material}

\subsection*{Experimental Setup} \noindent In our experiments, given $K$ resource categories, we simulate the resource capacity of each category by enforcing an even spacing principle if possible such that $l_k=l_0+k\Delta L$ for $k \in \{1, 2, \hdots K-1\}$, where $l_0$ is the location of the first exit, $\Delta L=\nfloor{\frac{N-l_0}{K-1}}$ and $l_K=N$ is the last exit, i.e. the full model. We perform experiments with ResNet56~\cite{resnet} on CIFAR-10 and with DenseNet121~\cite{densenet} on CIFAR-100. We use the default ResNet settings for 56-layer architecture and insert two evenly spaced early exits at the 18th and 36th layers (K=3). For DenseNet, we follow the default settings for 121-layer configuration and insert three early exit layers at the 12th, 36th and 84th layers due to transition layers (K=4). We train these models using Adam optimizer~\cite{adam} for 150 epochs (first 20 epochs without early exits) and a batch size of 128, with the initial learning rate of 0.1 (decays by 0.1 at 50th and 100th epochs). On the ImageNet dataset, we use MSDNet~\cite{msdnet} with 35 layers, 4 scales and 32 initial hidden dimensions. We insert four evenly spaced early exits at the 7th, 14th, 21th and 28th layers (K=5). Each early exit classifier consists of three 3x3 convolutional layers with ReLU activations. We set $\alpha_{KL}=0.01$ and activate it after completing 75\% of the training. For HRNet, we inject exits after the 2nd and 3rd stages with structures as in~\cite{seg_adp}. For BERT, we insert three evenly spaced early exits at the 3rd, 6th and 9th layers (K=4). Each early exit classifier consists of a fully-connected layer. We finetune the models for 20 epochs using gradient descent with a learning rate of 3e-5 and batch size of 16.

For our approach, based on validation performance, we set $D_h=0.5D$ and $D_h=2D$ for image/text classification experiments respectively. For image segmentation, we first downsample predictions by four with bilinear sampling and then operate on the mean of pixel-level computations. We optimize the weights using Adam optimizer with the learning rate of 3e-5 on validation data and set $\alpha_{cost}=10$. We observe that our algorithm satisfies the budget constraint with this setting and it is also robust with respect to selections within the range of $\alpha_{cost} \in [1, 100]$. In all experiments, we stop the optimization if the loss does not decrease for 50 consecutive epochs on the validation set. Inference measurements for CIFAR experiments are carried out on a machine with an 8-core 2.9GHz CPU, other experiments on a machine with RTX3060 GPU, and repeated ten times. The extra inference time caused by the exit score computations is also included in the reported latency measurements, and the cost is much smaller compared to the cost of the forward pass of the model as shown in Table \ref{tab:model_stats}.

\subsection*{Effect of Self-Distillation}
\noindent
We use the same model trained with self-distillation while comparing our scheduling policy with other early exit methodologies in all reported results. To analyze the improvements obtained with self-distillation, we also report the results on CIFAR datasets without applying self-distillation during training in Table~\ref{tab:results_dist}. Compared to the results provided in Table~\ref{tab:results_cv}, we observe up to ~1\% accuracy decrease in EENet and BranchyNet, and up to ~1.6\% accuracy decrease in MSDNet when self-distillation during training is disabled.

\begin{table}[h!]
\centering
\resizebox{\columnwidth}{!}{%
\begin{tabular}{c|c|ccc}
{ \textbf{Dataset}}& { \textbf{Budget}} & { \textbf{BranchyNet}} & { \textbf{MSDNet}} & { \textbf{EENet}} \\ \hline
 & { 3.50 ms}& { 93.55}& { 93.60}  & { 93.69} \\
 & { 3.00 ms}& { 92.11}& { 92.39}  & { 92.58} \\
\multirow{-3}{*}{{ CIFAR-10}}  & { 2.50 ms}& { 86.98}& { 88.10}  & { 88.51} \\ \hline
 & { 7.50 ms}& { 73.90}& { 73.88}  & { 74.05} \\
 & { 6.75 ms}& { 70.99}& { 70.96}  & { 71.54} \\
\multirow{-3}{*}{{ CIFAR-100}} & { 6.00 ms}& { 67.09}& { 67.15}  & { 68.56}
\end{tabular}} 
\caption{\small Accuracy (\%) values under various budget settings on CIFAR datasets without self-distillation during multi-exit model training.}\label{tab:results_dist}
\end{table}

\begin{figure}[b]
    \centering
    \includegraphics[width=\columnwidth]{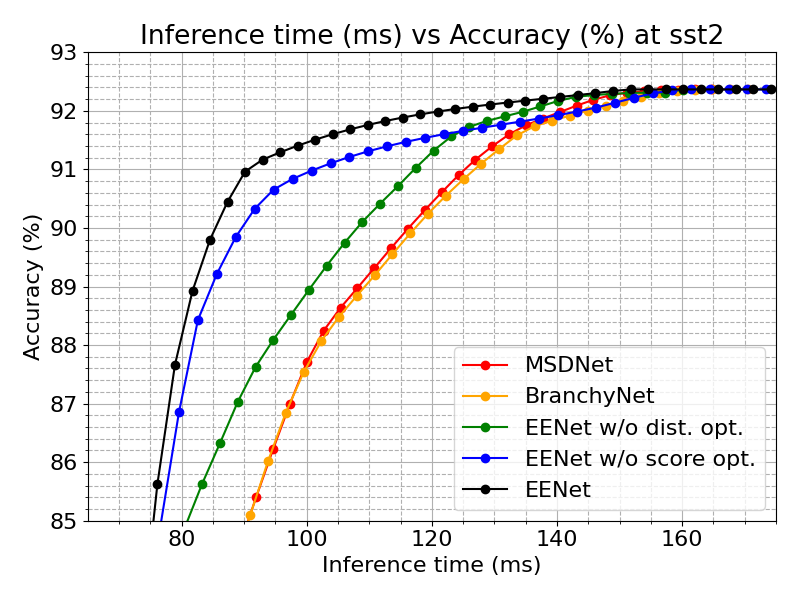}
    \caption{\small Average latency (ms) vs Accuracy (\%) results at SST-2 for BranchyNet, MSDNet, and EENet variations (without distribution/scoring optimization).}
    \label{fig:test}
\end{figure}

\begin{figure*}[t!]
    \centering
    \includegraphics[width=\textwidth]{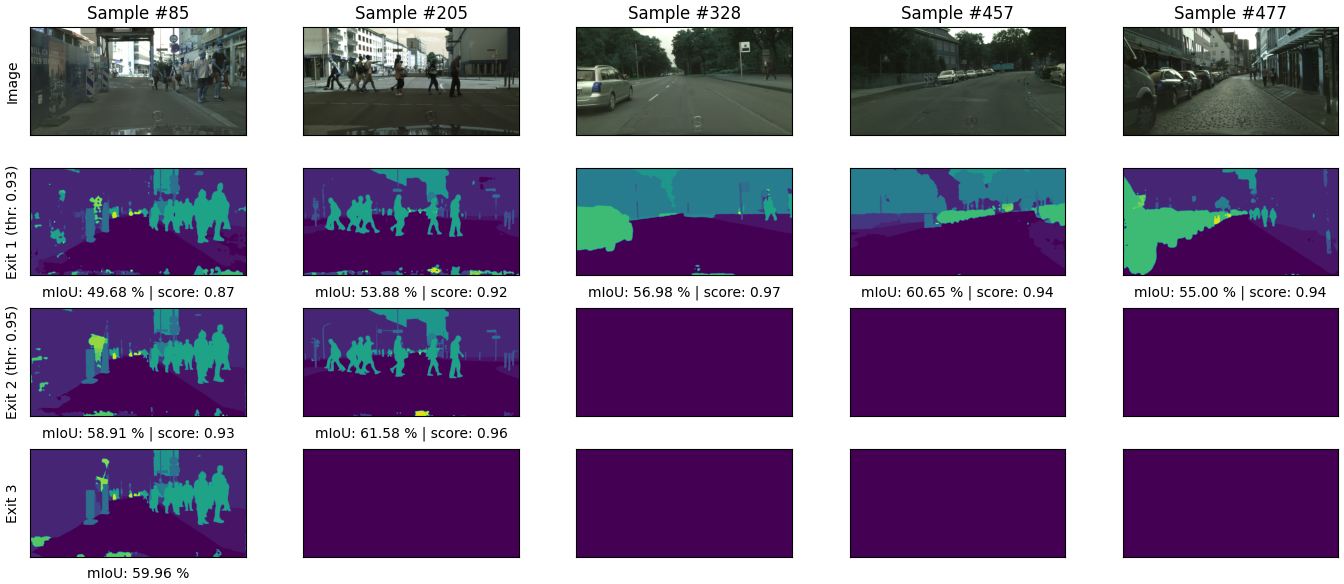}
    \caption{\small Visual comparison of randomly selected image segmentation examples from Cityscapes dataset for 50 ms/sample budget. Empty images indicate that the sample has exited previously as a result of having a higher exit score than the computed exit threshold.}
    \label{fig:seg}
\end{figure*}

\subsection*{Ablation Study of Design Components}
\noindent We also analyze the effect of different components in EENet on performance by in-depth investigation of the results for SST-2. Figure \ref{fig:test} provides the plot of average inference time vs. accuracy for two additional variants of EENet and compares with MSDNet and BranchyNet. The first variant of our approach shows the results of our system without optimizing the exit scoring, and instead, directly using maximum prediction scores. The second variant shows the results of EENet without optimizing exit distributions through our budget-constrained learning, and instead, directly using geometric distribution. We observe that optimization of both exit scoring functions and distributions to obtain thresholds contributes to the superior performance of EENet.

\subsection*{Visual Analysis of Early Exit Behavior}
\begin{figure*}
    \centering
    \begin{subfigure}[b]{0.22\textwidth}
        \centering
        \includegraphics[width=\textwidth]{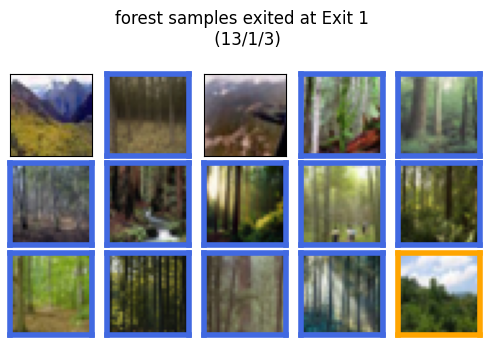}
        \caption[Network2]%
        {{\small Exit 1 - forest}}    
    \end{subfigure}
    \begin{subfigure}[b]{0.22\textwidth}  
        \centering 
        \includegraphics[width=\textwidth]{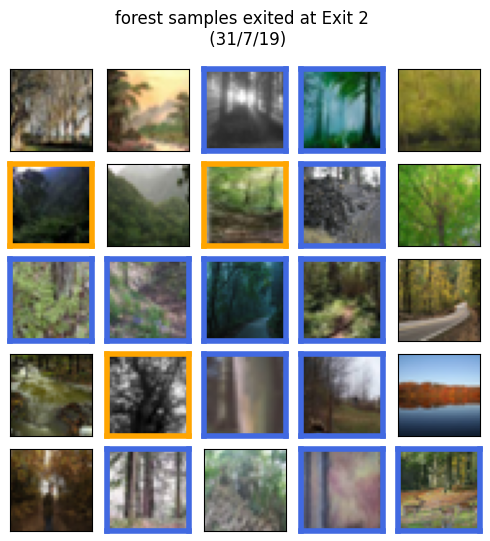}
        \caption[]%
        {{\small Exit 2 - forest}}    
    \end{subfigure}
    \begin{subfigure}[b]{0.22\textwidth}   
        \centering 
        \includegraphics[width=\textwidth]{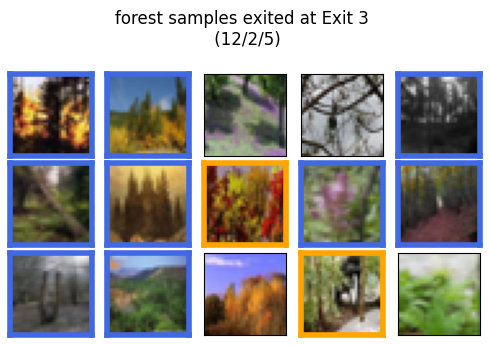}
        \caption[]%
        {{\small Exit 3 - forest}}    
    \end{subfigure}
    \begin{subfigure}[b]{0.25\textwidth}   
        \centering 
        \includegraphics[width=\textwidth]{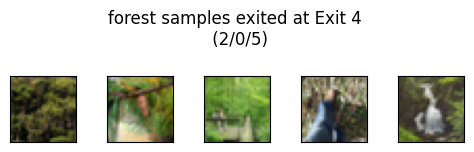}
        \caption[]%
        {{\small Exit 4 - forest}}    
    \end{subfigure}
    \vspace{-7.5pt}
    \vskip\baselineskip
        \begin{subfigure}[b]{0.22\textwidth}
        \centering
        \includegraphics[width=\textwidth]{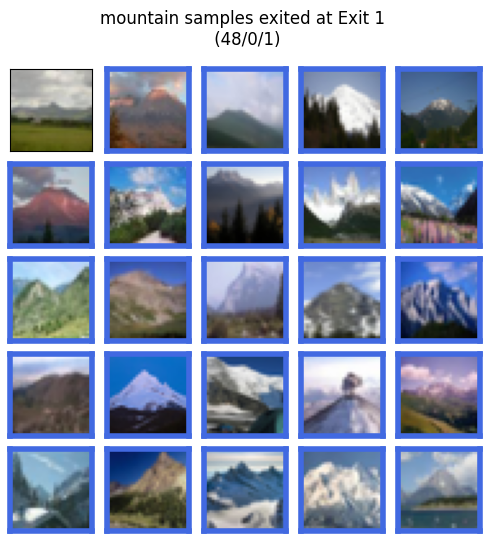}
        \caption[Network2]%
        {{\small Exit 1 - mountain}}    
    \end{subfigure}
    \begin{subfigure}[b]{0.22\textwidth}  
        \centering 
        \includegraphics[width=\textwidth]{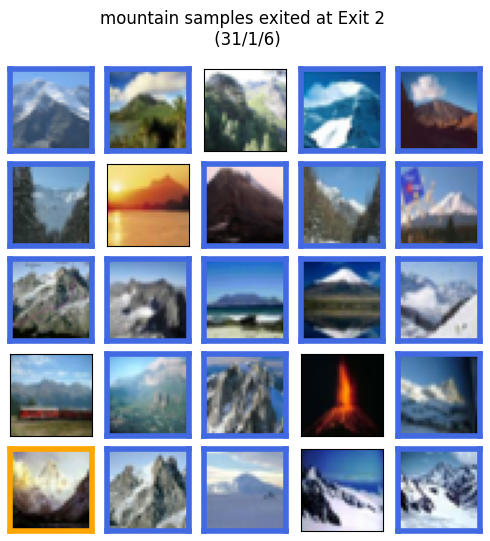}
        \caption[]%
        {{\small Exit 2 - mountain}}    
    \end{subfigure}
    \begin{subfigure}[b]{0.24\textwidth}   
        \centering 
        \includegraphics[width=\textwidth]{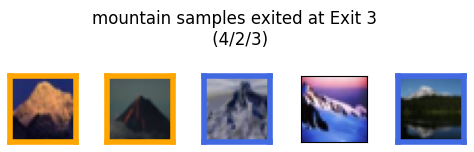}
        \caption[]%
        {{\small Exit 3 - mountain}}    
    \end{subfigure}
    \begin{subfigure}[b]{0.20\textwidth}   
        \centering 
        \includegraphics[width=\textwidth]{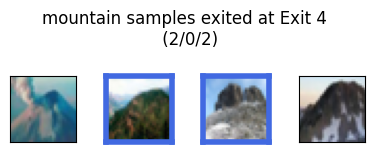}
        \caption[]%
        {{\small Exit 4 - mountain}}    
    \end{subfigure}
    \vspace{-7.5pt}
    \vskip\baselineskip
        \begin{subfigure}[b]{0.22\textwidth}
        \centering
        \includegraphics[width=\textwidth]{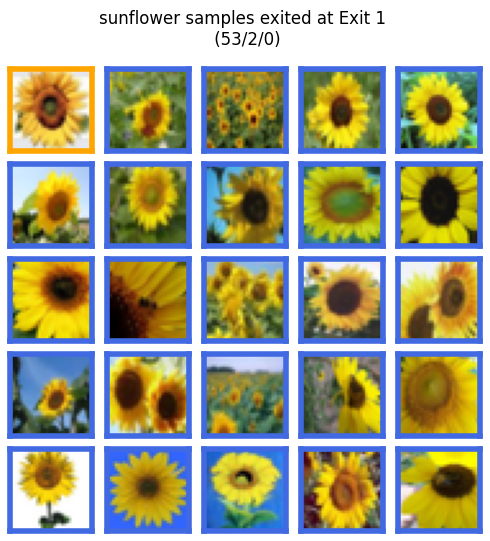}
        \caption[Network2]%
        {{\small Exit 1 - sunflower}}    
    \end{subfigure}
    \begin{subfigure}[b]{0.22\textwidth}  
        \centering 
        \includegraphics[width=\textwidth]{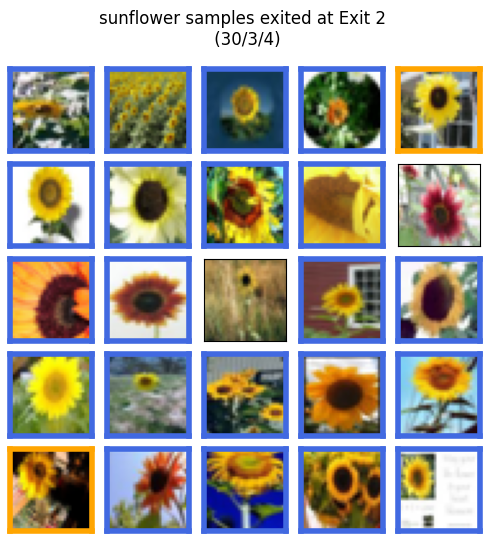}
        \caption[]%
        {{\small Exit 2 - sunflower}}    
    \end{subfigure}
    \begin{subfigure}[b]{0.25\textwidth}   
        \centering 
        \includegraphics[width=\textwidth]{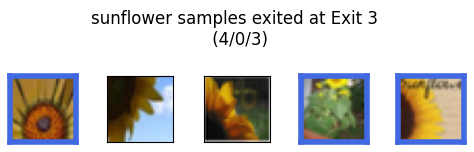}
        \caption[]%
        {{\small Exit 3 - sunflower}}    
    \end{subfigure}
    \begin{subfigure}[b]{0.2\textwidth}   
        \centering 
        \includegraphics[width=\textwidth]{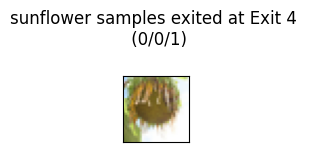}
        \caption[]%
        {{\small Exit 4 - sunflower}}    
    \end{subfigure}
    \vspace{-7.5pt}
    \vskip\baselineskip
        \begin{subfigure}[b]{0.20\textwidth}
        \centering
        \includegraphics[width=\textwidth]{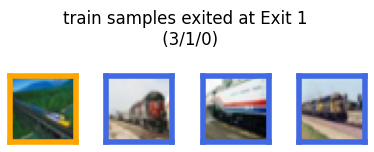}
        \caption[Network2]%
        {{\small Exit 1 - train}}    
    \end{subfigure}
    \begin{subfigure}[b]{0.22\textwidth}  
        \centering 
        \includegraphics[width=\textwidth]{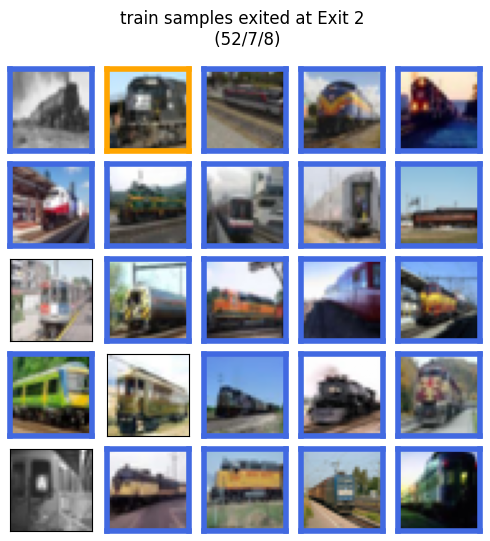}
        \caption[]%
        {{\small Exit 2 - train}}    
    \end{subfigure}
    \begin{subfigure}[b]{0.22\textwidth}   
        \centering 
        \includegraphics[width=\textwidth]{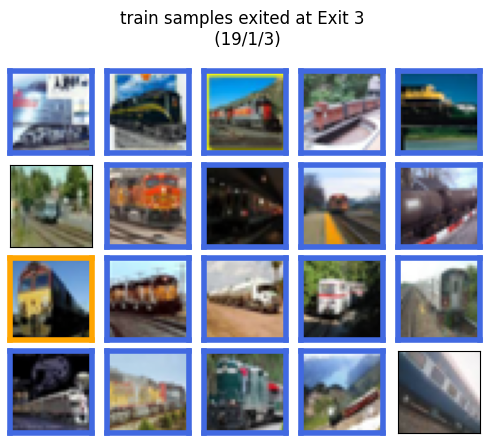}
        \caption[]%
        {{\small Exit 3 - train}}    
    \end{subfigure}
    \begin{subfigure}[b]{0.26\textwidth}   
        \centering 
        \includegraphics[width=\textwidth]{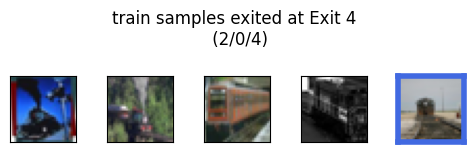}
        \caption[]%
        {{\small Exit 4 - train}}    
    \end{subfigure}
    \caption{ Samples from CIFAR-100 test set for forest/mountain/sunflower/train classes. Each subfigure illustrates the samples at the corresponding exit of DenseNet121 with four exits under the average inference budget of 6.25 milliseconds/sample. Blue borders indicate correct predictions. Orange borders indicate incorrectly predicted samples and if it were exiting at the last exit, it would be correctly predicted. No borders indicate the samples incorrectly predicted by EENet at that exit and also the last exit. The values on the titles indicate the number of samples from these categories (correctly predicted / incorrectly predicted at that exit / also incorrectly predicted at the last exit).}
    \label{fig:example}
\end{figure*}

\begin{figure*}[b!]
    \includegraphics[width=\textwidth]{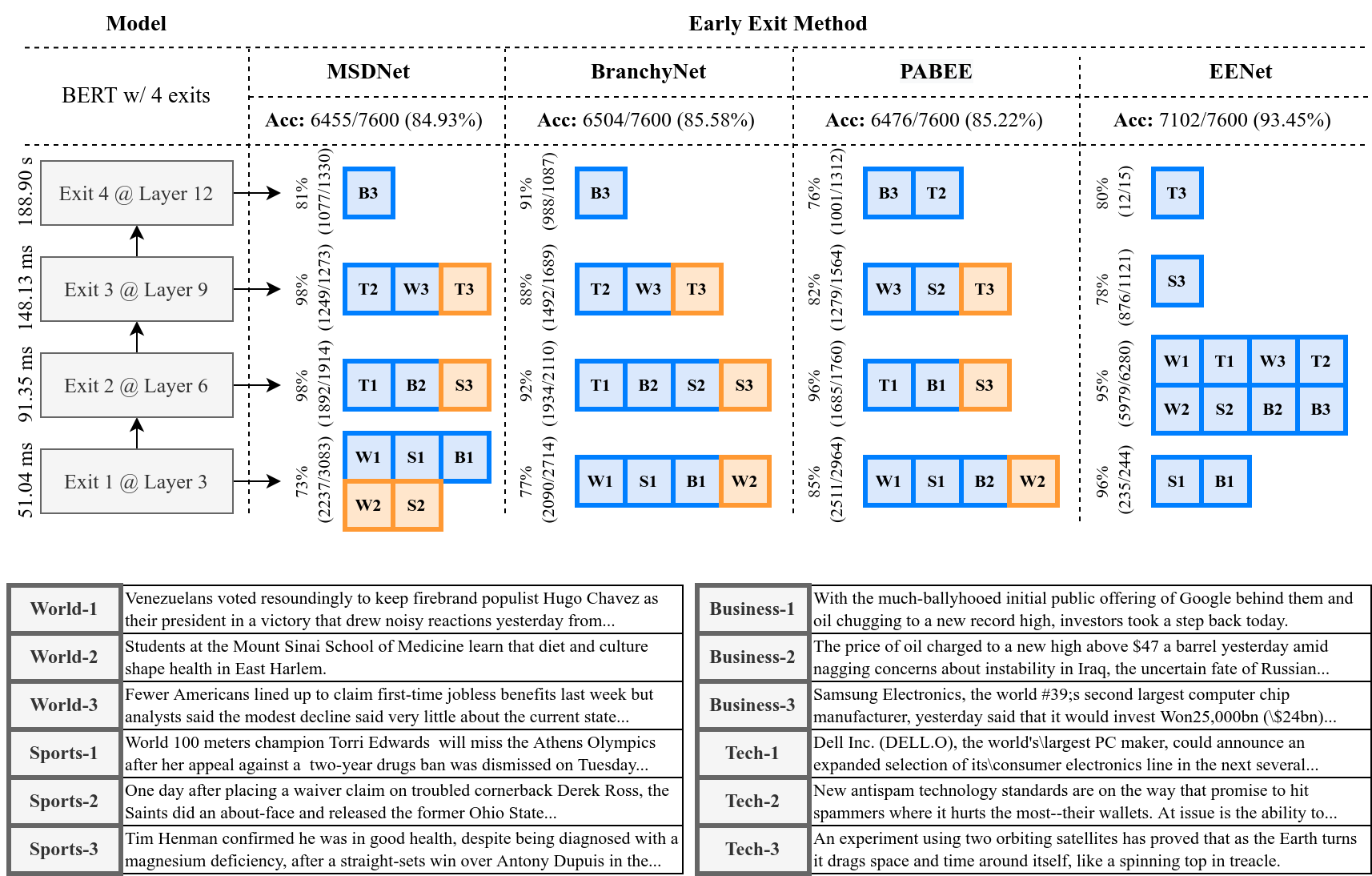}
    \caption{\small Visual comparison of the early exit approaches on AgNews test data with BERT (4 exits) for the average latency budget of 100 milliseconds. We illustrate the randomly selected twelve samples from four classes and the exit location that they were assigned. Images with green/red borders are predicted correctly/incorrectly at the corresponding exit. We also report the number of correct predictions and exited samples at each exit. Our approach does not make costly assignments to the last two exits and uses the second exit more effectively.}
    \label{fig:comp2}
\end{figure*}

\noindent We analyze the early exiting behavior of EENet on different tasks by qualitatively investigating the samples. On image segmentation, we observe that frames with less objects tend to exit earlier as shown in Figure~\ref{fig:seg}. We also illustrate test samples exited at each exit for four different classes from CIFAR-100 data in Figure~\ref{fig:example}. It is visually clear that the easier samples exit earlier to utilize the provided more efficiently. We observe that EENet utilizes the second exit in this particular scenario very efficiently by assigning easy samples and obtaining higher accuracy with significantly lower average latency. We conduct a similar analysis for the experiment set on AgNews test data in Figure~\ref{fig:comp2}. Similarly, EENet utilizes the second exit efficiently by assigning easy samples and obtaining higher accuracy with significantly lower latency.

\subsection*{Adapting to Dynamic Budget Settings} 
\noindent In practice, the test dataset may contain significantly easier/harder or out-of-distribution samples. Data distribution can also change over time. However, none of the existing adaptive inference algorithms strictly meets the latency budget over test data since the optimization is performed over the training or validation dataset. We also optimize the scheduling policy over the validation dataset however, our solution provides lightweight easy-to-optimize schedulers without requiring any changes on the full model itself. Therefore, a simple yet effective approach of switching between a few schedulers (optimized for different budget values) in an online manner is possible in our framework and can be utilized if necessary. For instance, we consider the scenario for CIFAR experiments, where during the test, we can switch between schedulers trained for three different budget values as provided in Table~\ref{tab:results_cv}. For the 3.0 ms/sample budget setting on CIFAR-10, if the test samples are easier/harder than expected and the realized latency per sample is getting lower/higher than the provided budget, we can switch to the scheduler optimized for the budget of 3.50/2.50 ms per sample. To this end, we compute the remaining budget per sample after each inference operation and switch to the scheduler trained under the closest budget setting. We provide the results in Table~\ref{tab:results_online}.

\begin{table}[]
\centering
\resizebox{\columnwidth}{!}{%
\begin{tabular}{c|c|l|c|c}
{ \textbf{Dataset}} & { \textbf{Budget}} & \textbf{Method}        & { \textbf{Latency}} & { \textbf{Accuracy (\%)}} \\ \hline
{ }                                         & { }                                        & BranchyNet             & { 2.87 ms}                                  & { 92.57}                                          \\
{ }                                         & { }                                        & EENet                  & { 2.85 ms}                                  & { 92.90}                                          \\
\multirow{-3}{*}{{ CIFAR-10}}               & \multirow{-3}{*}{{ 3.00 ms}}               & EENet w/ online switch & { 2.98 ms}                                  & { 92.92}                                          \\ \hline
{ }                                         & { }                                        & BranchyNet             & { 6.55 ms}                                  & { 71.65}                                          \\
{ }                                         & { }                                        & EENet                  & \multicolumn{1}{c|}{6.61 ms}                                     & 72.12                                                                 \\
\multirow{-3}{*}{{ CIFAR-100}}              & \multirow{-3}{*}{{ 6.75 ms}}               & EENet w/ online switch & \multicolumn{1}{c|}{6.74 ms}                                     & 72.11                                                             
\end{tabular}}
\caption{\small Test accuracy values on CIFAR datasets and the realized latency/sample values during the test for BranchyNet, EENet, and EENet with online switching during inference.} \label{tab:results_online}
\end{table}

\end{document}